# Multiple-Kernel Based Vehicle Tracking Using 3D Deformable Model and Camera Self-Calibration

Zheng Tang [1], Gaoang Wang [1], Tao Liu [2], Young-Gun Lee [1], Adwin Jahn [1], Xu Liu [3], Xiaodong He [4], Jenq-Neng Hwang [1]

[1]Department of Electrical Engineering
University of Washington
Seattle, WA, USA
{zhtang, gaoang, lygstj, adwin555, hwang}@uw.edu

[2]School of Information and Communication Engineering
Beijing University of Posts and Telecommunications
Beijing, China
tao.liu@bupt.edu.cn

[3]Institute of Industrial Science
University of Tokyo
Tokyo, Japan
liuxu@kmj.iis.u-tokyo.ac.jp

[4]Deep Learning Technology Center
Microsoft Research
Redmond, WA, USA
xiaohe@microsoft.com

*Abstract*—Tracking of multiple objects is an important application in AI City geared towards solving salient problems related to safety and congestion in an urban environment. Frequent occlusion in traffic surveillance has been a major problem in this research field. In this challenge, we propose a model-based vehicle localization method, which builds a kernel at each patch of the 3D deformable vehicle model and associates them with constraints in 3D space. The proposed method utilizes shape fitness evaluation besides color information to track vehicle objects robustly and efficiently. To build 3D car models in a fully unsupervised manner, we also implement evolutionary camera self-calibration from tracking of walking humans to automatically compute camera parameters. Additionally, the segmented foreground masks which are crucial to 3D modeling and camera self-calibration are adaptively refined by multiple-kernel feedback from tracking. For object detection/ classification, the state-of-the-art single shot multibox detector (SSD) is adopted to train and test on the NVIDIA AI City Dataset. To improve the accuracy on categories with only few objects, like bus, bicycle and motorcycle, we also employ the pretrained model from YOLO9000 with multi-scale testing. We combine the results from SSD and YOLO9000 based on ensemble learning. Experiments show that our proposed tracking system outperforms both state-of-the-art of tracking by segmentation and tracking by detection.

## I. INTRODUCTION

Thanks to the rapid growth of hardware performance and the amount of visual data in recent years, computer vision on intelligent surveillance systems has attracted more and more attention, especially the application in AI City. In traffic surveillance, multiple object tracking (MOT) is a crucial field that can be applied in many tasks, including traffic flow calculation, safe driving, etc. Currently, MOT16 [1] has been the benchmark dataset where most of the objects are human beings and the scenarios are much different from vehicle tracking in traffic surveillance. In the NVIDIA AI City Dataset, frequent occlusion by other vehicles, trees, and lighting posts has been a major problem for robust tracking of vehicle objects. In [2][3][4], our research group proposes constrained multiple-kernel (CMK) tracking to address the occlusion problem. Multiple kernels are used to represent several parts of each object, so that when one or some of the kernels are occluded, we can put larger weights to other visible kernels and link all the kernels based on some predefined constraints. However, for vehicle objects in traffic videos, the occluded parts are usually not regular due to the viewing perspectives caused by the fast car movement, so we make use of 3D deformable models of vehicles to define multiple kernels in 3D space [5][6]. Another benefit from building 3D vehicle models is that we could understand the vehicle attributes while tracking, including vehicle type, speed, orientation, etc, so as to improve the detection performance. It is also possible to automatically locate the regions of license plates for re-identification [7].

Besides video frames, 3D car modeling also requires camera parameters and segmented foreground masks as input. In AI City, due to the huge number of cameras within the network, it is unrealistic to manually calibrate each camera. Our research group proposes to utilize camera self-calibration [8] from tracking of detected human objects in the field of view (FOV) to automatically compute the projection matrix. The noise in the computation of vanishing points is handled by mean shift clustering and Laplace linear regression through convex optimization. The estimation of distribution algorithm (EDA), an evolutionary optimization scheme, is then used to optimize the locations of vanishing points and the estimated camera parameters, so that all the unknown camera parameters can be fine-tuned simultaneously. For robust object segmentation, we use the Multi-kernel Adaptive Segmentation and Tracking (MAST) system presented by us [9][10][11]. More specifically, after preliminary segmentation and tracking in each frame, we dynamically compute the penalty weights for the thresholds in background subtraction and shadow detection based on two multiple-kernel feedback loops to preserve foreground in regions with similar color to the background.

In the stage of object detection/classification, the state-of-the-art Single Shot MultiBox Detector (SSD) [12] is adopted in our experiments on the NVIDIA AI City Dataset. SSD has been evaluated on many benchmark datasets, e.g., PASCAL [13] and COCO [14], which have confirmed its competitive accuracy and computation efficiency in object detection. Since some classes like bus, motorcycle, pedestrian and bicycle have very few instances in the training dataset, we also adopt the pre-trained model from YOLO9000 [15] to help detect such categories. Moreover, a multi-scale testing method is applied in the testing stage to detect far-away small objects. Then we combine the detection results from SSD and YOLO9000 according to ensemble learning.

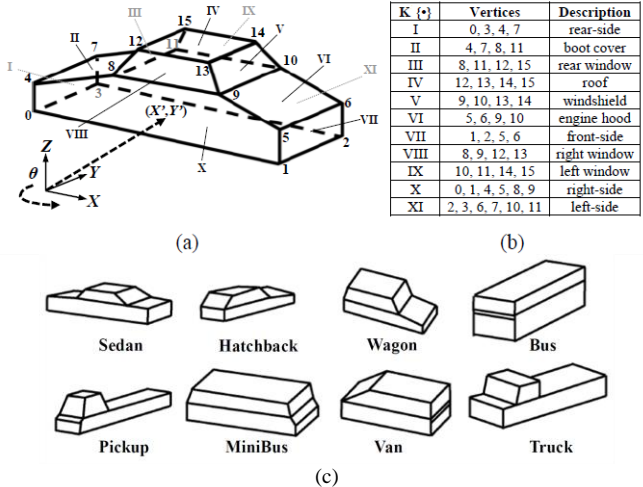
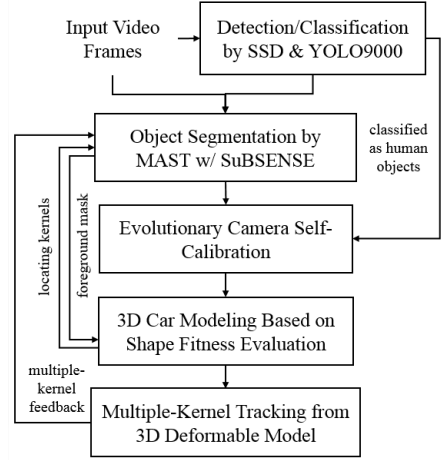

Fig. 1. (a) Generic model for 3D vehicle modeling [20]. (b) Table of the kernels built from the 3D vehicle model. (c) Different types of vehicle deformed from generic model.

Fig. 2. Overview proposed system framework.

The rest of this paper is organized as follows. In Section II, we give a brief review of other works in MOT and 3D vehicle modeling. Our proposed framework is detailed in Section III. Section IV presents the experimental results and analyses. Finally, we draw the conclusion in Section V.

## II. RELATED WORK

### A. Multiple Object Tracking

Most of the top-ranked methods on the benchmark of MOT [1] depend on object detection for target initialization, which are classified in the school of tracking by detection. One representative method is simple online and real-time tracking (SORT) [16] that is based on rudimentary data association and state estimation techniques to produce object identities on the fly. Recently, Milan et al. propose the first online MOT algorithm based on deep learning [17] that achieves top performance on the benchmarks.

Another category of MOT is tracking by segmentation. In [4], Chu et al. from our research group develop human tracking by combining CMK and adaptive Kalman filtering based on object segmentation. However, when the color of background is similar to parts of the objects, failure in segmentation will lead to error in tracking as well, i.e., the problem of object merging. To address this, we propose the MAST framework [8] to refine the segmented foreground blobs by multiple-kernel feedback. MAST has achieved the best single-camera tracking accuracy on the benchmark dataset, NLPR_MCT [18].

The methods in the above two categories are not specifically designed for vehicle tracking in traffic videos, and thus they can easily fail when complex occlusion happens in our scenarios.

Kernel-based object tracking, e.g., mean shift tracker [19], has been widely used for tracking a single target, because of its fast convergence and low computation. To address the problem of occlusion, Chu et al. improve single-kernel tracking into CMK tracking [2][3]. The researchers in our lab continue to extend the CMK work for vehicle tracking in 3D space, where multiple kernels are built in 2D from the 3D vehicle model. The constraints include constant 3D distance and relative yaw and pitch angles between kernels [5][6].

### B. 3D Vehicle Modeling

Zhang et al. [20] first propose to generate approximate 3D vehicle model deformed from a 3D generic model. They assume that the camera is static and well calibrated, i.e., the 3×4 projection matrix $\mathbf{P}$ is known. The 3D deformable model is constructed by 16 vertices and 23 arcs as shown in Fig. 1(a). The vehicle shape is defined by 12 shape parameters, including the vehicle length, vehicle widths, vehicle heights, etc. The vehicle pose is determined by 3 parameters, which are its position on the ground plane and its orientation about the vertical axis perpendicular to the ground plane. These 15 parameters can be estimated by evaluating the fitness of 3D deformable model, which is conducted in an evolutionary way based on EDA. The fitness evaluation score (FES) is used as the objective function for evolutionary computation. FES is defined as the sum of gradient magnitudes of pixels along the perpendicular direction of each projected line segments of the 3D vehicle model.

## III. METHODOLOGY

The overview of our proposed architecture for tracking of multiple vehicle objects based on 3D deformable model and camera self-calibration is depicted in Fig. 2. Each module is detailed in the following subsections.

### A. Object Detection/Classification

First, we train our SSD framework based on training dataset from NVIDIA. Moreover, a pre-trained model from YOLO9000 is adopted to help detect categories with few training instances like bus, motorcycle, pedestrian and bicycle. To better detect far-away small objects, we use multi-scale testing strategy. For each testing frame, the image is divided into 9 sub-regions with 50% overlapping between each pair of neighboring sub-regions. Then the individual results from each sub-region are combined together according to non-maximum suppression.

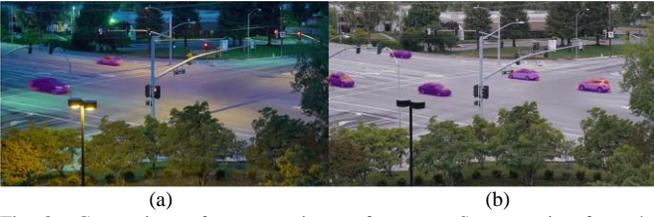

Fig. 3. Comparison of segmentation performance. Segmentation from the preliminary result of SuBSENSE with shadow detection is depicted in blue and segmentation after the application of multiple-kernel feedback loops is shown in red. (a) Example frame from the experimental video "walsh_santomas_20170603_016.mp4" (b) Experimental frame from the experimental video "walsh_santomas_20170603_019.mp4".

To adaptively select and combine the results from SSD and YOLO9000, we propose a scheme of ensemble learning. First, the intersection over union (IOU) ratios of the detected bounding boxes from SSD and YOLO9000 are calculated. If an IOU ratio is higher than a threshold (empirically set as 0.5 ), it is assumed that the two detectors locate the same object. If the predictions are of the same class, a linear regression is employed to merge the two detected bounding boxes into one,

$$\hat{B} = w_1 B_1 \oplus w_2 B_2, \quad (1)$$

where $B_1$ and $B_2$ are the two detected bounding boxes from SSD and YOLO9000, respectively; $\hat{B}$ is the ground truth of bounding box and $w_1$ and $w_2$ are the training parameters. $B_1$, $B_2$ and $\hat{B}$ are all vectors of 4 dimensions, each consisting of center coordinate, width and height of the corresponding bounding box. On the other hand, if the two predictions from SSD and YOLO9000 are different in categories, the following regression is used to determine which classification we should trust,

$$\hat{y} = w_3 s_1 + w_4 s_2, \quad (2)$$

where $s_1$ and $s_2$ are confidence scores of detection from SSD and YOLO9000 respectively; $\hat{y} = 1$ if the prediction from SSD is correct and $\hat{y} = -1$ if the prediction from YOLO9000 is correct; $w_3$ and $w_4$ are training parameters.

*B. Multiple-Kernel Adaptive Segmentation*

Segmented foreground blobs of human objects are used in camera self-calibration and the tracking system to predict poses of objects. In the traffic videos, most vehicle and human objects share similar color and/or chromaticity with background, which causes poor performance of segmentation, i.e., the problem of object merging. To ensure robust estimation of camera parameters and build accurate 3D car model, we implement a multiple-kernel feedback scheme to adaptively control the thresholds in segmentation. More specifically, according to the feedback from tracking, two sets of multiple kernels are built in the current frame and modeled background respectively within the object region. One set of kernel histograms are created in YCbCr color space, and the other only use the Cb and Cr channels. Comparing the Bhattacharyya distance of kernel histograms between the current frame and the background, we can measure their color/chromaticity similarity. A fuzzy Gaussian penalty weighting function is designed to map the similarity proportionally to a penalty weight, $w_{\text{pen}}$, to be added on thresholds in background subtraction and shadow detection:

$$w_{\text{pen}} = \begin{cases} \exp\left[-\frac{9 \cdot (1.0 - simi)^2}{4 \cdot (1.0 - simi_{\min})^2}\right], & simi_{\min} \leq simi < simi_{\max} \\ 0, & \text{otherwise} \end{cases}, \quad (3)$$

where $simi$ is the color/chromaticity similarity computed from the corresponding set of kernel histograms. The $simi_{\min}$ and $simi_{\max}$ indicate the region of $simi$ value to perform re-segmentation. The penalization on threshold values is defined by multiplying $(1 - w_{\text{pen}})$. Meanwhile, since the preliminary foreground blob may fail to cover the entire object body, the kernel region to conduct re-segmentation is expanded by a factor of $w_{\text{pen}}/2$. In our current realization, the presented MAST scheme is combined with the state-of-the-art change detection algorithm, SuBSENSE [21], where a shadow detection module is added. A comparison of segmentation before and after applying multiple-kernel feedback is shown in Fig. 3.

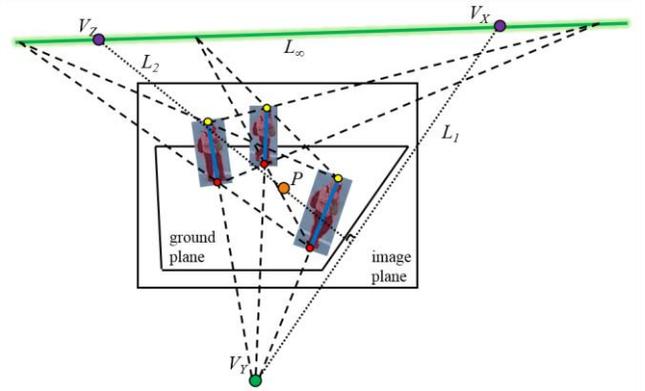

Fig. 4. Geometry of estimation of vanishing points in self-calibration (ideal scenario). The three blue poles represent three positions of the same walking person appearing at different frames, with yellow and red endpoints indicating their head and foot points respectively. Point $V_Y$ in green marks the vertical vanishing point, and the green line $L_\infty$ denotes the horizon line, which is constructed based on the head-head and foot-foot line intersection of all pairs of the same walking person. The dashed lines are auxiliary lines for the search of $V_Y$ and $L_\infty$. Points $V_X$ and $V_Z$ in purple are the other two vanishing points located on $L_\infty$. Point $P$ in orange gives the principal point of the camera. The dotted lines $L_1$ and $L_2$ are auxiliary lines for locating $V_Z$.

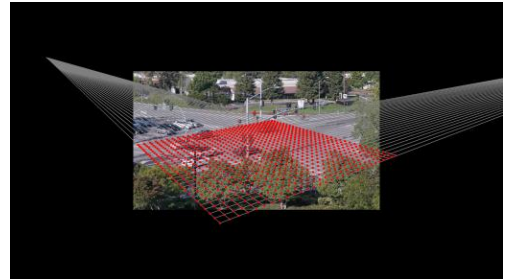

Fig. 5. Modeled ground plane built from the camera parameters estimated from evolutionary camera self-calibration. The red dots form a 30 meter × 30 meter 3D grid on the ground plane projected to 2D space.

*C. Evolutionary Camera Self-Calibration*

The proposed camera self-calibration framework mainly depends on reliable human body segmentation and EDA to search for optimal locations of vanishing points and optimize the camera parameters, so that we can exploit the availability of human tracking and segmentation data for robust calibration.

First, the head and foot points for each tracking position of detected pedestrians are located based on the results generated from MAST. In other words, each human position is modeled as a pole vertical to the ground plane, which is extracted as the major axis of the segmented foreground blob. Ideally, when there is neither noise nor outlier, the vertical vanishing point, $V_Y$, and the horizon line, $L_\infty$, can be easily determined as illustrated in Fig. 4. However, due to the existence of noise and outliers, this scenario is unrealistic in real world. There are always many inconsistent candidate points of $V_Y$ each generated by a pair of tracking positions. Similarly, many candidate points of $L_\infty$ may not lie on the same line as well. Therefore, we propose to use mean shift clustering and Laplace linear regression for noise reduction in the estimation of $V_Y$ and $L_\infty$. In [22], Caprile and Torre introduce the method to recover both intrinsic and extrinsic parameters from given vanishing points, $V_X$ and $V_Z$, according to some assumptions on intrinsic camera parameters. To further reduce uncertainty in computation caused by inevitable noise, the locations of $V_X$ and $V_Z$ are optimized based on minimization of the standard deviation of estimated 3D human heights. Then, the camera parameters are also optimized according to minimization of the reprojection error on the ground plane, where the assumptions on intrinsic camera parameters can be relaxed at the same time. The Estimation of Multivariate Normal Algorithm – global (EMNA$_{global}$) [23], a type of multivariate EDA, is adopted for both optimizations. In Fig. 5, we present the modeled ground plane in experimental videos which is estimated from our derived camera parameters.

### D. CMK Tracking Based on 3D Vehicle Modeling

In the dataset of traffic videos, occlusion is a major problem leading to failure in tracking. This can be overcome by CMK tracking, whose main concept is to emphasize the visible parts (kernels) of an object while maintaining their relative spatial constraints. However, the allocation of kernels within an object region is another difficulty that we need to concern. Thanks to 3D vehicle modeling, the problem can be easily resolved by regarding each patch/plane of a 3D vehicle model as a kernel, which is demonstrated in Fig. 1.

In the proposed vehicle tracking system, we combine the Kalman-filtering framework with 3D vehicle model-based CMK tracking. The segmented foreground masks from MAST are first input to the system. Then, Kalman prediction is conducted based on the segmented objects. If the 3D vehicle model is not built or needs to be updated, the predicted pose is used to construct 3D deformable model; otherwise, the CMK tracker makes use of the pre-built 3D vehicle model to track the object. The final tracking results are used to update the Kalman filter. The Kalman-filtering framework also enables our system to handle total occlusion in short periods.

The cost function of 3D CMK tracking is given by

$$J(x) = \sum_{\kappa=1}^{n_\kappa} w_\kappa \left( J_\kappa^s(x) + J_\kappa^f(x) \right), \quad (4)$$

where $x \in \mathbb{R}^3$ is the location of each kernel, $w_k$ is an adjustable weight, and $\kappa = 1, 2, \ldots, n_\kappa$ are the indices of kernels. $J_\kappa^s(x)$ is the color similarity term and $J_\kappa^f(x)$ is the fitness term.

To compute $J_\kappa^s(x)$, for each visible kernel $\mathbf{K}\{\kappa\}$, the color information is associated by back-projecting each 2D point $p^\kappa$ in the kernel region to its 3D position $\overline{P}^\kappa$ on the kernel plane. Therefore, the target kernel $\mathbf{K}\{\kappa\}$ can be described by its probability density function $q$ in terms of the $r$-bin histograms,

$$q_u^\kappa = \frac{\sum_i \sum_{\kappa=1}^{n_\kappa} k\left(\left\|\frac{P_c^\kappa - \overline{P}_i^\kappa}{h}\right\|^2\right) \delta[b(p_i^\kappa) - u]}{\sum_i \sum_{\kappa=1}^{n_\kappa} k\left(\left\|\frac{P_c^\kappa - \overline{P}_i^\kappa}{h}\right\|^2\right)}, \quad \sum_{u=1}^r q_u^\kappa = 1, \quad (5)$$

where $\|\cdot\|$ denotes the L2 norm, the subscript $c$ represents the center of mass of the kernel, the subscript $i$ gives each pixel location inside the kernel, $h$ is the bandwidth of 3D kernel, $\delta$ is the Kronecker delta function and $k(\cdot)$ is a Gaussian kernel function for weighting in space. The function $b$ associates the pixel at location $p_i^\kappa$ with the index of its bin in the color histogram. During CMK tracking, all the kernels search for the regions with the highest similarity to the target kernel, where $J_\kappa^s(x)$ is inversely proportional to the similarity function. The fitness term $J_\kappa^f(x)$ is defined similarly except the Kronecker delta function is swapped by the total FES of 3D vehicle model.

To describe spatial constraints between 3D kernels, we first choose a reference kernel $\mathbf{K}\{\kappa^*\}$ which has the maximum visible area. The first constraint is that the distance between each kernel and the reference kernel should remain the same as $L_{\kappa,\kappa^*}$, which implies

$$\left\|P_c^\kappa - P_c^{\kappa^*}\right\|^2 = L_{\kappa,\kappa^*}^2, \text{ for any visible } \mathbf{K}\{\kappa | \kappa \neq \kappa^*\}, \quad (6)$$

Second, the pitch $\phi_{\kappa,\kappa^*}$ and yaw $\varsigma_{\kappa,\kappa^*}$ between $\mathbf{K}\{\kappa^*\}$ and $\mathbf{K}\{\kappa | \kappa \neq \kappa^*\}$ should be the same as well. The two vectors $v_a$ and $v_b$ which are orthogonal to each other and cross $P_c^{\kappa^*}$ are calculated by

$$v_a = \frac{P_a^\kappa + P_o^\kappa}{2} - P_c^{\kappa^*}, \quad v_b = \frac{P_b^\kappa + P_o^\kappa}{2} - P_c^{\kappa^*}, \quad (7)$$

where $P_o^\kappa$ is the intersection of two adjacent line segments selected from $\mathbf{K}\{\kappa\}$. $P_a^\kappa$ and $P_b^\kappa$ are the end points of both line segments respectively. Let us define $v_{\kappa,\kappa^*} = P_c^\kappa - P_c^{\kappa^*}$. The second constraint is obtained by

$$\begin{cases} \frac{v_a \cdot v_{\kappa,\kappa^*}}{\|v_a\| \|v_{\kappa,\kappa^*}\|} = \cos(\phi_{\kappa,\kappa^*}) \\ \frac{v_b \cdot v_{\kappa,\kappa^*}}{\|v_b\| \|v_{\kappa,\kappa^*}\|} = \cos(\varsigma_{\kappa,\kappa^*}) \end{cases}, \text{ for any visible } \mathbf{K}\{\kappa | \kappa \neq \kappa^*\}. \quad (8)$$

The projected gradient method [2][3] is adopted to iteratively solve this constrained optimization problem efficiently.

To further improve the fitness of 3D vehicle model during vehicle orientation, we first predict the turning angle by Kalman filtering and then select the model with the highest FES within a range of 10 degrees. Moreover, when there is no occlusion detected but FES is lower than a certain threshold, the 3D vehicle model will be updated.

### IV. EXPERIMENTAL RESULTS

The NVIDIA AI City Dataset consists of video data sources recorded by cameras aimed at intersections in urban areas in diverse conditions covering both daytime and nighttime. There

TABLE I
EXPERIMENTAL RESULTS OF TRACK 1

| Categories | aic480 | aic540 | aic1080 |
|---|---|---|---|
| Car | 0.75 | 0.61 | 0.59 |
| SUV | 0.52 | 0.48 | 0.45 |
| Van | 0.22 | 0.24 | 0.22 |
| Bicycle | 0.38 | 0.05 | 0.03 |
| Bus | 0.35 | 0.46 | 0.45 |
| Motorcycle | 0.14 | 0.29 | 0.22 |
| Pedestrian | 0.00 | 0.03 | 0.10 |
| GroupOfPeople | -- | 0.12 | 0.09 |
| Signal-R | -- | 0.00 | 0.38 |
| Signal-G | -- | 0.00 | 0.30 |
| Signal-Y | -- | 0.00 | 0.06 |
| S-Truck | 0.45 | 0.48 | 0.45 |
| M-Truck | 0.19 | 0.27 | 0.27 |
| L-Truck | 0.02 | 0.14 | 0.14 |
| mAP | 0.34 | 0.25 | 0.28 |

TABLE II
COMPARISON OF DETECTION OF RARE-INSTANCE CLASSES ON *AIC1080*

| Methods | Bus | Bicycle | Motorcycle | mAP |
|---|---|---|---|---|
| SSD | 0.214 | 0.034 | 0.149 | 0.267 |
| SSD+YOLO9000 | 0.304 | 0.035 | 0.164 | 0.274 |
| SSD+YOLO9000 w/ MST | 0.336 | 0.040 | 0.168 | 0.277 |

TABLE III
COMPARISON OF PERFORMANCE OF MULTIPLE OBJECT TRACKING

| Methods | MOTA% | MOTP% | FAF | FP | FN | ID Sw. |
|---|---|---|---|---|---|---|
| *CMK3D* | **82.0** | **99.5** | **0.23** | **7** | *310* | **0** |
| *MAST* | *79.8* | 91.9 | *0.26* | 118 | **214** | 23 |
| *Kalman* | 64.2 | 86.4 | 0.46 | 197 | 404 | 29 |
| *RNN* | 69.0 | 96.3 | 0.40 | 53 | 484 | *8* |
| *SORT* | 61.8 | *99.1* | 0.50 | *13* | 629 | 30 |

Note that red-bold entries indicate the best results in the corresponding columns for each video sequence, and blue italics the second-best.

are more than 80 hours of videos in total with 1080p or 720x480 resolution. Collaboratively, all the teams contribute over 1.4M annotations of 14 different classes.

*A. Object Detection/Classification*

In Track 1 of the NVIDIA AI City Challenge, our team adopts the VGG-based SSD network to train on the joint datasets of *aic480* and *aic540*. The network is based on a pre-trained model on ImageNet. We set the number of iterations as 200,000 with a batch size of 16. A TensorFlow re-implementation of the original Caffe code is used for this challenge. We use 512 inputs instead of 300 to enhance detection of small objects like traffic lights.

Our submitted method based on the combination of SSD and YOLO9000 with multi-scale testing ranks 4[th] in terms of overall mAP among the 18 finalist teams. The details of the results are shown in the Table I. The complete results of Track 1 can be found on the challenge website: http://smart-city-conference.com/AICityChallenge/results.html. Since *aic540* and *aic1080* are from the same sources with different resolutions, we only compute testing results on *aic540* and linearly scale up for *aic1080*. Interestingly, the overall mAP of *aic1080* improves by 0.03 compared with *aic540*. It is because the computation of average precision in this challenge ignores objects smaller than 30x30 pixels, and thus the mAP improvement shows that multi-scale testing is capable of detecting small objects in *aic1080* which are filtered out in *aic540*. Table II further demonstrates the effectiveness of multi-scale YOLO by comparing mAP and AP for classes with rare instances. With the same parameter settings as SSD + YOLO9000, multi-scale strategy creates leading one of the results in *aic1080* validation dataset.

*B. Multiple Object Tracking*

Our experimental data are from two videos within the sub-dataset of Silicon Valley Intersection. We manually label 1,760 tracking locations as ground truth including 32 objects across 1,356 frames. The proposed method labeled as *CMK3D* is compared with four different tracking algorithms, including two based on tracking by segmentation, *MAST* [9] and *Kalman* [4], and the other two in tracking by detection, *RNN* [17] and *SORT* [16]. Note that *mast* is currently the state-of-the-art on the NLPR_MCT benchmark [18] and *RNN* achieves one of the top performances on the MOT16 benchmark [1].

The comparison of experimental results can be seen from Table III. The standard metrics used in MOT Challenge [1] are adopted to evaluate each method. Multiple Object Tracking Accuracy (MOTA) measures tracking accuracy combining three error sources: false positives, missed targets and identity switches. Multiple Object Tracking Precision (MOTP) measures the misalignment between the annotated and the predicted bounding boxes. FAF represents the average number of false alarms per frame. FP, FN, and ID Sw. stand for the total numbers of false positives, false negatives (missed targets) and identity switches respectively. The proposed method achieves the best performance in all metrics except for FN. It is because *MAST* [9] is designed for preserving more foreground around the object regions for robust tracking by segmentation. However, extra background information may be included in adaptive segmentation which causes the increase of FP and ID Sw. The capability of *CMK3D* in resolving occlusion can be learned from the fact that there is no identity switch, while all the other approaches tend to generate new object identities when occlusions happen. The state-of-the-art tracking-by-detection approach *RNN* can recover most identities after short periods of occlusion, however, cannot continuously track objects in regions under occlusion, resulting in low MOTA.

Another benefit of the proposed method is that tracking locations are not only in 2D space, but also can be back-projected to the 3D ground plane using the projection matrix automatically computed from evolutionary camera self-calibration. Therefore, real vehicle attributes such as speed and orientation angle can be estimated accurately. Moreover, the constructed 3D vehicle model is available during tracking, which can be used to interpret vehicle type, color, etc. We can even locate the license plates of each vehicle based on its 3D model and perform further re-identification if the resolution is sufficiently high. Some examples of 3D deformable models for different types of vehicles built for tracking are shown in Fig. 6.

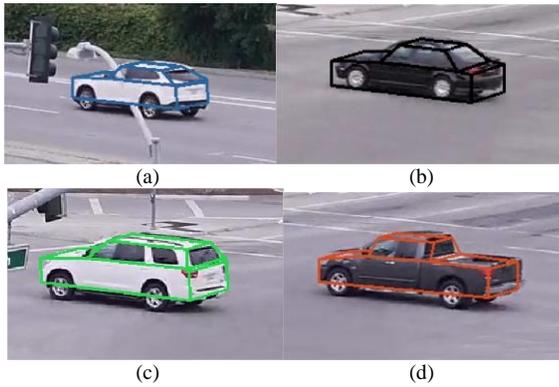

Fig. 6. Examples of 3D deformable models for different types of vehicles built for CMK tracking. (a) SUV. (b) Sedan. (c) Van. (d) Small truck.

Our team is selected as the winner of Track 2 for the value and innovation of our proposed approach, along with the success of our demonstration. The demo videos can be watched on our website: http://allison.ee.washington.edu/thomas/3dvt/.

## V. Conclusion

In Track 1, the combined efforts of SSD and YOLO9000 with multi-scale testing are employed based on ensemble learning, which facilitates us in the detection of categories with few objects. Our overall mAP ranks at the $4^{th}$ place.

In Track 2, we propose a fully unsupervised 3D vehicle tracking framework assisted by camera self-calibration. It is capable of overcoming frequent occlusion in the NVIDIA AI City Dataset. Experiments show that the proposed method outperforms both state-of-the-art of tracking by segmentation and tracking by detection.

In the future, we plan to improve the accuracy of object detection/classification by considering feedback of vehicle types from 3D vehicle modeling. This method can also be extended to tracking/re-identification across multiple cameras.

## Acknowledgment

The 3D CMK vehicle tracking framework is developed based on the implementation by the honorable graduates from the Information Processing Lab, Dr. Kuan-Hui Lee, who is currently a research scientist at the Toyota Research Institute, and Dr. Chun-Te Chu, who is currently a software development engineer at Microsoft.